**Title:** Beyond hospital reach: Autonomous lightweight ultrasound robot for liver sonography

**Authors:**
Zihan Li[1]†, Yixiao Xu[1]†, Lei Zhang[2], Taiyu Han[3], Xinshan Yang[1], Yingni Wang[1], Mingxuan Liu[1], Shenghai Xin[4], Linxun Liu[4], Hongen Liao[1]*, Guochen Ning[1,2]*

**Affiliations:**
[1]School of Biomedical Engineering, Tsinghua University; Beijing, P.R. China.
[2] School of Clinical Medicine (Beijing Tsinghua Changgung Hospital); Beijing, P.R. China.
[3]College of Applied Sciences, Shenzhen University; Shenzhen, Guangdong, P.R. China.
[4]Qinghai Provincial Peoples Hospital; Xining, P.R. China.

*Corresponding author. Email: liao@tsinghua.edu.cn (H.L.)
ningguochen@tsinghua.edu.cn (G.N.)
†These authors contributed equally to this work

**Abstract:** Liver disease is a major global health burden. While ultrasound is the first-line diagnostic tool, liver sonography requires locating multiple non-continuous planes from positions where target structures are often not visible, for biometric assessment and lesion detection, requiring significant expertise. However, expert sonographers are severely scarce in resource-limited regions. Here, we develop an autonomous lightweight ultrasound robot comprising an AI agent that integrates multi-modal perception with memory attention for localization of unseen target structures, and a 588-gram 6-degrees-of-freedom cable-driven robot. By mounting on the abdomen, the system enhances robustness against motion. Our robot can autonomously acquire expert-level standard liver ultrasound planes and detect pathology in patients, including two from Xining, a 2261-meter-altitude city with limited medical resources. Our system performs effectively on rapid-motion individuals and in wilderness environments. This work represents the first demonstration of autonomous sonography across multiple challenging scenarios, potentially transforming access to expert-level diagnostics in underserved regions.

**One-Sentence Summary:** The lightweight robot enables autonomous liver non-continuous standard plane sonography across multiple scenarios.

**Main Text:**

## INTRODUCTION

Liver disease represents a major global health burden, accounting for over two million deaths annually—approximately 4% of worldwide mortality. Cirrhosis and hepatocellular carcinoma constitute the predominant causes of liver-related fatalities. Meanwhile, parasitic infections pose additional challenges, particularly in resource-limited settings (*1-3*). Echinococcosis exemplifies this burden as a globally distributed zoonotic disease, with an estimated 2-3 million people infected annually, resulting in approximately 19300 deaths worldwide (*4*). The prolonged asymptomatic nature of many liver diseases complicates early diagnosis and necessitates long-term surveillance with timely intervention for optimal patient outcomes (*5, 6*). Given these diagnostic challenges, abdominal ultrasonography has been regarded as the primary imaging modality for liver assessment due to its non-invasive nature, widespread accessibility, and cost-effectiveness. Current clinical guidelines consistently recommend ultrasound as the first-line diagnostic tool for multiple hepatic diseases, including hepatocellular carcinoma surveillance in cirrhotic patients (*7*), non-alcoholic fatty liver disease screening, cholelithiasis detection, and echinococcosis diagnosis and monitoring (*4, 8, 9*). To standardize diagnostic protocols and improve accuracy, clinical guidelines have established predefined sets of standard imaging planes that ensure comprehensive liver visualization and pathological assessment (*10*).

Artificial intelligence (AI) has demonstrated exceptional capabilities in ultrasound image analysis and diagnostic interpretation. Recent deep learning models have achieved superior diagnostic accuracy for liver lesions compared to experienced radiologists in multi-center studies (*11*), while specialized AI systems for hepatic echinococcosis diagnosis have shown exceptional performance across diverse clinical settings (*12*). These AI diagnostic tools demonstrate substantial potential for reducing physician workload and enhancing ultrasound diagnostics in regions with limited medical resources. However, the acquisition of high-quality diagnostic images remains critically dependent on operator expertise, creating a fundamental bottleneck before AI analysis (*13*). Even medically trained personnel, such as critical care paramedics, demonstrate significantly lower scanning success rates compared to experienced physicians (*14*), yet such specialized medical resources are often unavailable in rural clinics, remote areas, and underserved regions. Furthermore, sending expert sonographers to resource-limited areas presents additional challenges including logistical difficulties and extended travel times, which strains already limited healthcare resources (*15*).

Medical robots with a high level of autonomy have already demonstrated great potential to perform complex clinical tasks. (*16, 17*). Recent advances demonstrate this potential. Zhang et al.'s (*18*) AI co-pilot bronchoscope robot assists novice physicians in achieving precise operations , while Liu et al. (*19*) developed autonomous bronchoscopy robots for foreign body detection. In ultrasound robotics, autonomous systems show similar promise, with Su et al. (*20*) developing fully autonomous thyroid diagnosis systems and Jiang et al. (*21*) achieving autonomous carotid examination through imitation learning from expert teleoperation. However, a significant clinical gap exists in autonomous ultrasound research. Abdominal ultrasound constitutes the majority of clinical ultrasound examinations, with nearly 69% of all scans performed on abdominal organs, especially the liver (*22*). Yet, most autonomous ultrasound research focuses on superficial organs, such as vessels (*23*), thyroid (*24*) and spine (*25*), creating substantial disconnect between clinical demand and current research priorities.

Existing autonomous ultrasound robots employ either rule-based or learning-based approaches, yet both face significant limitations in abdominal sonography. Rule-based methods are effective for superficial organs with predictable anatomy, such as centering vessels in carotid scanning (*26, 27*). However, they are inadequate for abdominal imaging where the liver's position beneath the rib cage demands precise maneuvering to avoid rib shadowing, and substantial inter-subject variability in organ size, shape, and position precludes standardized scanning protocols. Reinforcement learning approaches face similar challenges, as they require realistic simulation environments for training, yet the combination of significant anatomical variability across individuals, respiratory-dependent organ positioning, and complex rib cage constraints makes accurate abdominal simulation intractable (*28*). While imitation learning has shown promise in probe force control and posture adjustment for superficial organs (*29, 30*), abdominal sonography presents unique challenges. Specifically, standard liver ultrasound planes are not spatially continuous and lack distinctive anatomical landmarks, resulting in complex, ambiguous images that are inherently difficult for automated systems to interpret compared to the well-defined structures found in superficial organ imaging. Moreover, current imitation learning systems for ultrasound acquires expert demonstration via unnatural interaction modes such as robotic arm dragging (*31*) or remote control (*32*) that differ from clinical practice, and conducting data collection and evaluation on identical robotic platforms limits cross-system generalizability.

Body-coupled robots offer distinct advantages over decoupled robots by maintaining consistent spatial relationships with anatomical targets and ensuring portability. In contrast, conventional ultrasound robots using industrial robotic systems compromise the inherent portability advantages of ultrasound technology. The coupling enables precise manipulation during patient movement and provides inherent safety through direct mechanical constraint. This coupling principle has proven effective across medical applications, such as Posselli et al.'s (*33, 34*) head-mounted surgical robot that eliminated respiratory motion artifacts and improved surgical outcomes on moving eyes. However, most autonomous ultrasound robotics relies on decoupled collaborative robots which introduces safety vulnerabilities during sensor failures or unexpected patient movement (*35*), including involuntary physiological motion and external disturbances such as ambulance transport. Moreover, these bulky robotic systems are inherently difficult to transport and deploy, contradicting the core portability advantage that makes ultrasound imaging widely accessible in clinical practice. Although Ning et al. (*36*) proposed an abdominal-mounted ultrasound system, it provides only four degrees of freedom (DoF) motion and lacks force sensing capabilities, making it inadequate for abdominal sonography that demands multi-axis manipulation and precise pressure control.

To address these gaps, we propose an autonomous lightweight abdominal-mounted ultrasound robot for multi-task liver sonography involving complex anatomical structures across diverse scenarios (Fig. 1). Our system comprises an AI agent and an ultrasound robot. Inspired by expert sonography practices, the agent fuses the features of ultrasound image, force and probe pose, and employs a memory attention mechanism to process sequential inputs, autonomously generating precise action trajectories for locating non-continuous standard imaging plane. Recognizing that ultrasound's inherent portability and accessibility are fundamental to its widespread clinical adoption, while conventional fixed-base robotic systems cannot meet these deployment requirements, we propose a novel 588-gram, six-degree-of-freedom and cost-efficient (less than 3000 USD) robot, which is mounted on the abdomen to overcome various sources of motion. We validated and demonstrated that our robot can acquire standard liver imaging planes autonomously with expert-level performance by comparing anatomical structure consistency between robotic and expert acquisitions in 11 unseen volunteers. Our robot was also tested on six patients with hepatic cysts, renal cysts, hemangioma, or fatty liver for pathological screening. Furthermore, we deployed our robot in Xining, a high-altitude city at 2261 meters above sea level with limited medical resources, for hepatic echinococcosis screening, demonstrating practical diagnostic capabilities in medically underserved regions. Our robot also exhibited robust performance on rapid-motion individuals and in wilderness where specialist physicians are unavailable, showcasing exceptional environmental adaptability. These experiments demonstrate that the autonomous capabilities, the superior coupling and the lightweight design enable more stable ultrasound image acquisition and expands the range of clinical applications compared to previous ultrasound robots. This successful integration of AI agent, clinical expertise and lightweight robot demonstrates the transformative potential of our cost-effective robot to democratize expert-level sonography, bridge healthcare gaps in resource-limited regions where specialist physicians are scarce or inaccessible.

# RESULTS

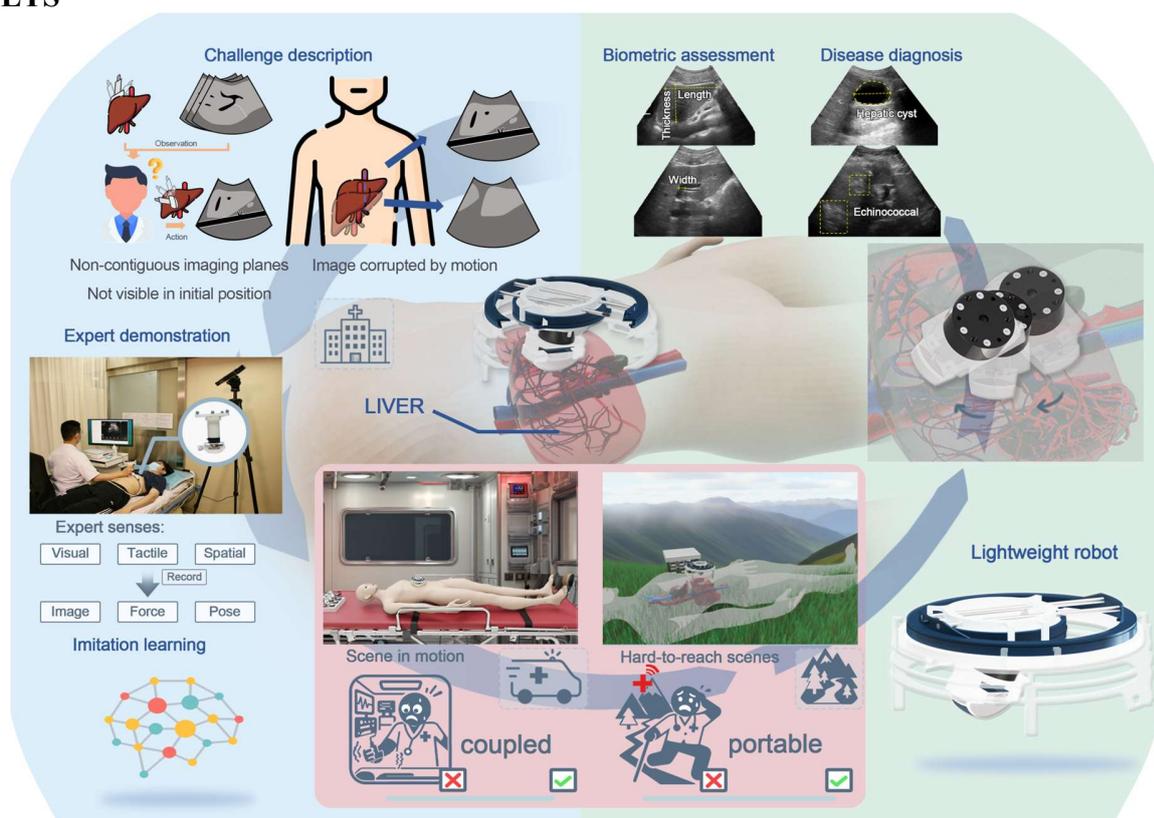

**Fig. 1. System overview.** Liver ultrasound examination encounters significant challenges including non-continuous standard imaging planes and motion that interfere with imaging decisions. We collect expert sonography demonstrations and transfer the clinical expertise to an AI agent through imitation learning. We deploy the learned clinical expertise to a lightweight abdominal-mounted robot to autonomously acquire standard imaging planes for disease diagnosis and biometric assessment. The tight human-robot coupling and portable design enable liver disease diagnosis in challenging environments such as dynamic scenarios, wilderness and high-altitude regions where specialized medical expertise is scarce.

## Expert demonstration collection

This research addresses abdominal sonography by modeling it as a context-dependent conditional generation process that integrates sequential multi-modal perceptual information. Deep abdominal sonography, particularly liver sonography, presents unique challenges compared to superficial organ examination. While surface organs can be localized using external visual landmarks, liver imaging relies entirely on ultrasound image for spatial navigation. The inherently two-dimensional nature of ultrasound images necessitates pose integration for spatial awareness, while the complex abdominal anatomy, containing both pressure-sensitive rigid structures (ribs) and deformable soft tissues (adipose layers), demands force sensing and control for safe and effective imaging. Therefore, our approach characterizes the ultrasound scanning process through three fundamental state modalities: ultrasound images, probe pose and contact forces, providing spatial, tactile, and visual information, respectively, with actions defined as target poses and forces. To capture these multi-modal demonstrations from expert sonographers, we developed a comprehensive data acquisition system (Fig. 1 expert demonstration) incorporating an optical tracking system for precise probe pose recording, a six-dimensional force sensor for contact force measurement, and a compact ultrasound device for real-time image acquisition. Unlike previous approaches, such as the remote-control systems or robotic arm manipulation, our system employs an ultrasound probe that closely mimics clinical probes, introducing no additional resistance or impedance, thereby ensuring that expert demonstrations authentically reflect clinical scanning procedures.

Four representative standard imaging planes were considered as evaluation targets and the corresponding demonstrations to locate these planes were recorded. These planes encompass core anatomical targets in routine abdominal ultrasonography, including (i) subcostal transabdominal aortic long-axis plane, displaying the complete abdominal aorta and left hepatic lobe, namely aorta plane, (ii) subcostal inferior vena cava (IVC) plane, visualizing the IVC long-axis with respiratory variation for volume status assessment, namely IVC plane, (iii) portal vein plane, demonstrating the portal vein sagittal section, round ligament of the liver, and inferior vena cava in a single transverse scanning plane, namely portal vein plane, and (iv) right intercostal hepatorenal plane, demonstrating the right kidney upper pole and hepatorenal plane, namely hepatorenal plane. These planes collectively encompass critical vascular, parenchymal, and potential pathological assessment zones essential for comprehensive abdominal evaluation.

**AI agent design and phantom experiment**

The agent employed a multi-modal memory module (M3 module) and a diffusion generator to map states to actions (Fig. 2A). The M3 module addressed two critical challenges in autonomous ultrasound examination as shown in Fig. 1. First, ultrasound examination requires inferring the non-continuous three-dimensional position of unseen targets from two-dimensional images, necessitating the integration of historical information for spatial reasoning. Second, due to human respiration and relative motion between human and robot, multiple ultrasound images with different contents are acquired at a single pose, introducing redundant and interfering information. To address these challenges, the M3 module adopted a sequential image-pose joint modeling approach (Fig. 2B). The module performed joint representation learning on historical ultrasound images and their corresponding poses, then incorporated an image importance scoring mechanism to filter out redundant information. This mechanism computed the sum of attention weights between each image and all positions as the importance score, selecting the $k$ ($k$=5) highest-scoring images and their corresponding poses in the memory for cross-attention fusion to generate integrated image-pose features. Meanwhile, since force is recorded in the task coordinate frame, the M3 module combined the selected $k$ poses with their corresponding force measurements, employing a Kolmogorov-Arnold Network (KAN) for force-pose feature fusion. The diffusion module is a conditional diffusion model that takes the resulting image-pose features and force-pose features as conditions to predict high-dimensional actions of $m$ ($m$=5) steps comprising six-dimensional poses and six-dimensional forces.

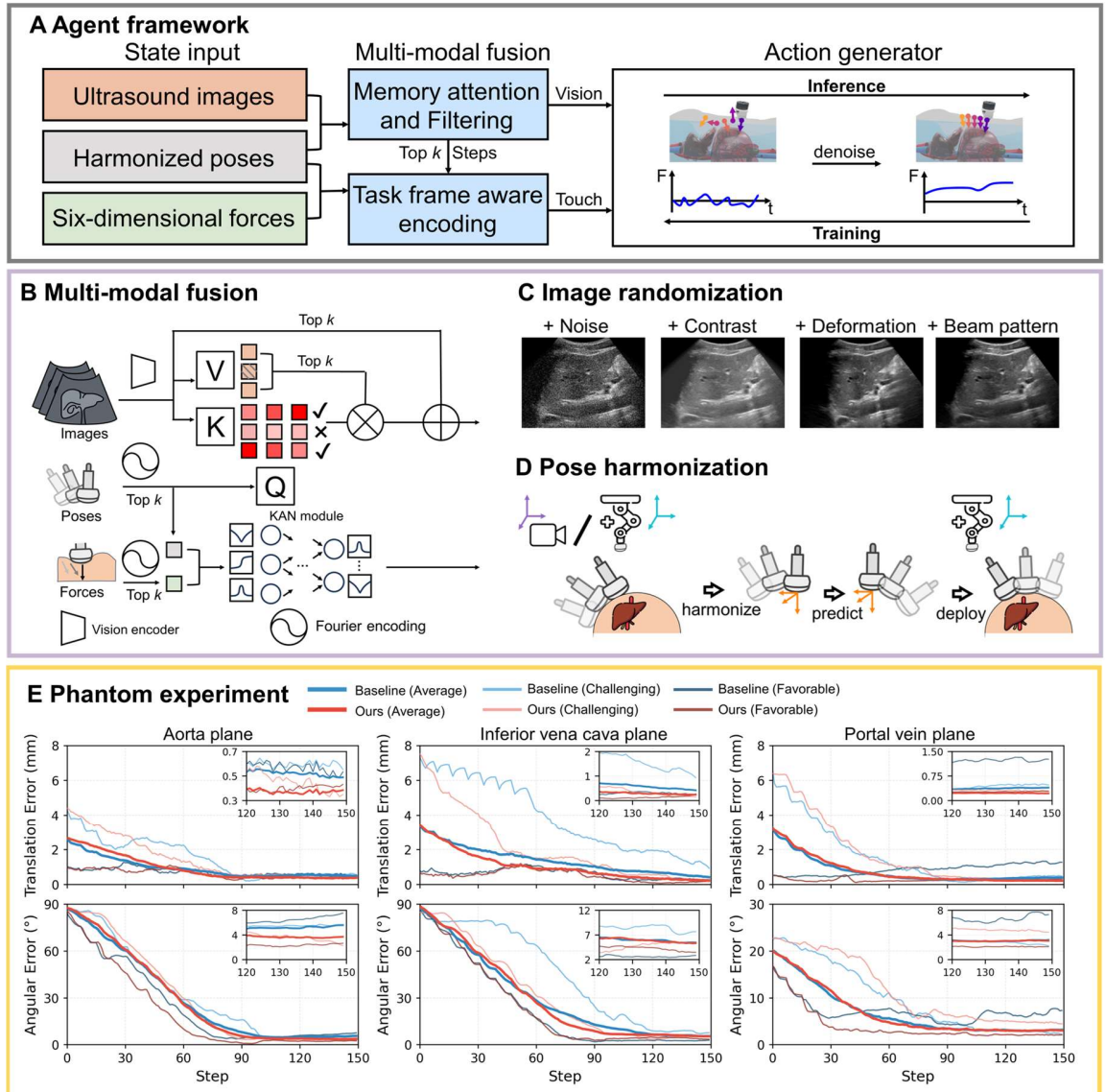

**Fig. 2. Imitation learning framework.** (**A**) Overall architecture. The agent utilizes a multi-modal fusion model to fuse the input sequential ultrasound images, harmonized poses, and six-dimensional force measurements through memory attention and select the most important k image-pose-force pairs, generating visual and tactile features for action generation. The action generator is a conditional denoising diffusion probabilistic model that predicts future trajectories of poses and forces. (**B**) Multi-modal fusion architecture. Images are encoded through a vision encoder, while poses and forces are processed through Fourier positional encoding. Cross-attention mechanisms fuse image and pose features, with the top $k$ most important features selected as the final visual features. Corresponding tactile features are generated from pose and force features through KAN (Kolmogorov-Arnold Network). (**C**) Image randomization. During training, the input images are randomly corrupted with noise, contrast adjustment, deformation, and beam pattern variations to enhance cross-subject generalization and robustness to imaging artifacts. (**D**) Pose harmonization. All poses are transformed to an ego-centric coordinate system relative to the current pose before network input. The predicted poses are subsequently transformed back to robot coordinates for execution. (**E**) Phantom

validation. Quantitative comparisons between baseline method (diffusion policy) and our approach across three standard planes, including aorta plane, inferior vena cava plane, and portal vein plane are shown. Mean translation errors and mean angular errors relative to the ground truth poses are plotted over execution steps for 14 independent trials per imaging plane. Our method is shown in red lines while baseline results are in blue. Representative challenging cases with large initial translation error (light colors) and favorable cases with smaller initial translation error (dark colors) are shown to illustrate the methods' performance across different difficulty levels.

To enhance the generalization capability of the agent across subjects and especially patients, we optimized both image and pose representations. At the image level, domain randomization was independently applied to each ultrasound image (Fig. 2C), introducing perturbations such as contrast variations and deformations that exhibit high inter-patient variability to strengthen the model's adaptability to individual differences. At the pose representation level, to eliminate the model's dependence on the spatial coordinate systems of specific demonstration or deployment devices, an ego-centric pose representation was adopted that transforms all poses into the relative coordinate system at the decision timepoint, achieving harmonized pose representation across different devices (Fig. 2D).

The proposed policy network was quantitatively evaluated on a liver phantom using a custom UR3e robot for AI agent validation before robotic system deployment. Three standard planes served as targets, including aorta plane, IVC plane, and portal vein plane, with ground truth poses recorded for each target. Fig. 2E shows translation errors and angular errors between the executed poses and ground truth poses during plane localization from 14 initial positions. Our policy efficiently reached target planes within 100 steps with faster convergence rates than the baseline diffusion policy (37) and maintained a relative stable pose once achieved. Compared to baseline, our approach demonstrated lower final translation errors and rotation errors. These results demonstrate enhanced trajectory efficiency and superior accuracy for initially unseen anatomical structures inference, providing a solid AI foundation for clinical deployment.

**Lightweight 6-DoF robot**

We designed an abdominal-mounted, planar 6-DoF cable-driven robot for ultrasound sonography. The 6-DoF design matches the kinematic capabilities of manual ultrasound examination. Inspired by cylindrical coordinate systems, horizontal translation was achieved through a single-plane configuration comprising one prismatic joint (Fig. 3C) and one revolute joint (Fig. 3D), while vertical translation was realized via a prismatic joint (Fig. 3E). The rotational degrees of freedom were implemented using a three-axis gimbal structure incorporating three revolute joints that enable pitch (Fig. 3F), yaw (Fig. 3G), and roll (Fig. 3H) rotations. A six-dimensional force sensor and a compact ultrasound probe were mounted at the gimbal's end-effector for force measurement and image acquisition, respectively.

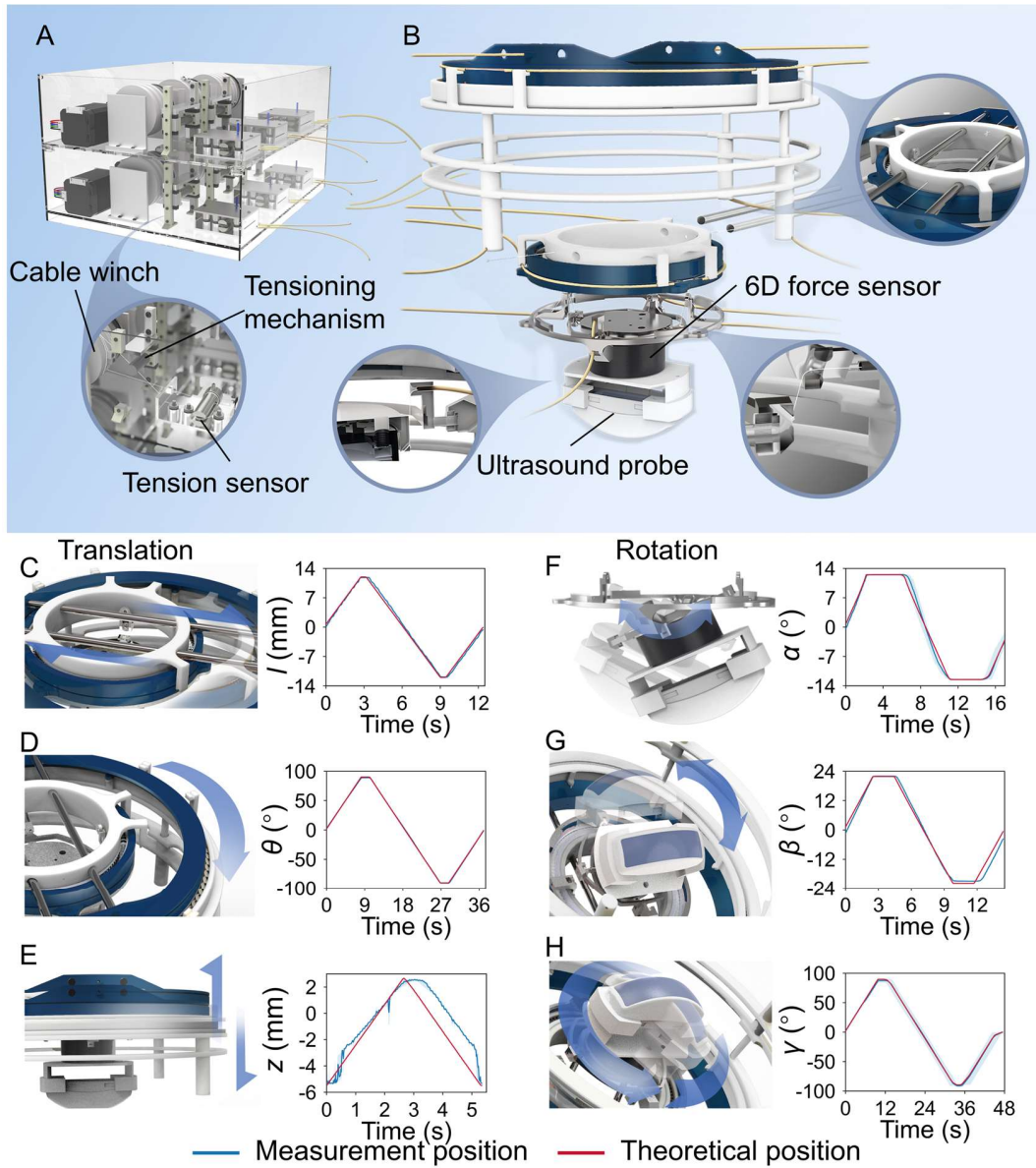

**Fig. 3. Abdominal-mounted, planar 6-DoF cable-driven robot.** (**A**) Cable-driven actuators. Each actuator comprises a cable winch that enables bidirectional cable operation (pulling and releasing) and a cable tension regulator consisting of a tensioning mechanism and tension sensor to maintain cable tautness. (**B**) End-effector. The end-effector provides six degrees of freedom, including translational motion (**C-E**) and rotational motion (**F-H**). Precision testing was conducted independently for each DoF ($n$=8). Theoretical mean values and standard deviations are represented by red lines and light red regions while measured mean values and standard deviations obtained via optical tracking are shown as blue lines and light blue regions. Motion ranges and test accuracies are listed in Table S1.

To improve portability, minimize patient burden and minimize system inertia during body mounting, we employed a cable-driven architecture that separates the actuators (Fig. 3A) from the joints. This design resulted in an ultrasound robot weighing 588 g, consisting of a 366.0 g end effector and 222.0 g probe-sensor assembly (Fig. S1). Each joint's motion was controlled through a dedicated cable winch and a pair of cables, forming a bidirectional closed-loop traction system with antagonistic "pull-release" operation for forward and reverse actuation. To address the inherent slack and backlash issues in cable-driven systems, we proposed a cable tension regulator. A tensioning mechanism maintained baseline tension in one cable, while a tension sensor integrated in series with the opposing cable employed threshold-based backlash elimination.

The robotic platform achieved high-precision and repeatable control performance. The curves in Fig. 3C-H illustrate the average trajectories for each degree of freedom ($n=8$) where the measurement curves and the theoretical curves were highly consistent. As shown in Table S1, the robotic system achieved high control accuracy. The cable tension regulator enhanced accuracy by up to 73.5% for DoF α. This high-precision robotic platform, with its clinical-equivalent DoFs and comprehensive hepatic workspace coverage, establishes a reliable foundation for autonomous sonography.

**Autonomous sonography on human**

The autonomous ultrasound robot was first deployed in the ultrasound department of Beijing Tsinghua Changgung Hospital to evaluate the feasibility of autonomous ultrasound scanning in clinical environments (Fig. 4A). The system performance was evaluated across three clinically relevant tasks: localizing the aorta plane (Fig. 4CD), IVC plane (Fig. 4EF), and portal vein plane (Fig. 4GH). For all tasks, untrained personnel mounted the robot on the volunteer's abdomen, initiating from randomized transverse imaging planes. During the experiments, subjects remained relaxed and breathed freely.

Our autonomous ultrasound robot demonstrated superior spatial reasoning capabilities during the systematic acquisition of abdominal imaging, mirroring the clinical scanning protocol employed by experienced sonographers. Taking the localization of the aorta plane as an example (Fig. 4CD), no aortic structure was visible in the initial ultrasound image (Fig. 4C(i)). To find the aorta structure, the robot autonomously executed a downward compression maneuver combined with a 1-centimeter displacement toward the patient's left side, consistent with the anatomical positioning of the abdominal aorta. Then, the short-axis view of the abdominal aorta was visible while liver boundary definition was enhanced, resulting in markedly increased image similarity scores (Fig. 4C(ii)).

Subsequently, the robot initiated the critical transition from short-axis to long-axis imaging through progressive yaw rotation, autonomously visualizing the longitudinal view of the abdominal aorta (Fig. 4C(iii)-C(iv)). During this phase, the robot accounted for the anatomical positioning of the abdominal aorta slightly left of the body midline and self-directed moderate leftward tilting angles (positive pitch values) to optimize the imaging plane orientation, demonstrating clinical-level anatomical awareness. To prevent excessive angulation that could result in suboptimal probe-rib contact, the robot autonomously executed rightward translational movements (negative Y-axis values), achieving optimal imaging quality through positional compensation rather than relying solely on angular adjustments.

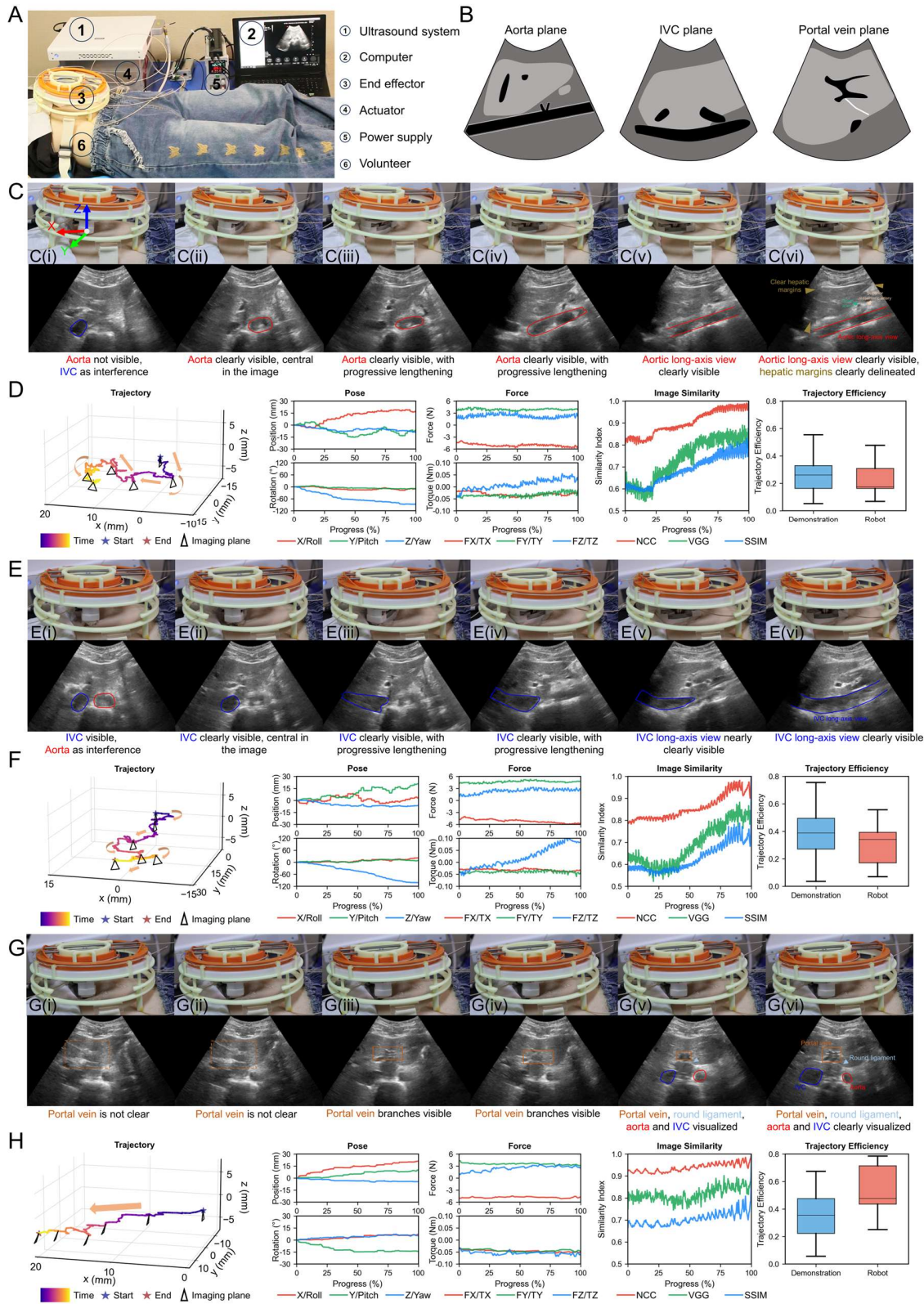

**Fig. 4. Autonomous standard plane localization process on previously unseen subject.** (**A**) Experimental setup demonstrating the robotic platform secured to the abdominal region via straps for stable human-robot coupling. (**B**) Schematic illustration of the targeted standard planes. (**C**, **F**, **I**) Sequential snapshots of robotic

motion, (**D**, **G**, **J**) corresponding ultrasound images during the autonomous localization of the three standard planes, including aortic plane, inferior vena cava (IVC) plane, and portal vein plane on a representative subject. Anatomical structures of interest are outlined in the ultrasound images with colored annotations matching the descriptive text below each image. (**E**, **H**, **K**) Quantitative analysis of autonomous sonography performance. Three-dimensional motion trajectories are plotted with the imaging planes represented as triangles and the coordinate system defined in figure C(i). Pose trajectories show three-dimensional translational and rotational movements during standard plane localizations. Force curves display contact forces and torques measured by the sensor throughout the sonography process. Image similarity progression tracks the computed similarity between acquired images and reference standard planes during autonomous localization. Trajectory efficiency was calculated as the ratio of total trajectory length to the distance between start and end points using 50 uniformly sampled waypoints of each trajectory. Comparison between expert demonstrations ($n$=134, 140, 184 for each plane) and autonomous robotic performance ($n$=11 subjects per plane) reveals comparable efficiency.

In the final phase, the robot autonomously refined its positioning as the yaw angle increased toward alignment between the probe's longitudinal axis and the body's sagittal plane, with the aortic long-axis structure progressively emerging with enhanced clarity. Meanwhile, the robot continued fine adjustments, ensuring that the final image not only clearly delineated the aortic long-axis structure but also simultaneously visualized the celiac and superior mesenteric arterial branches alongside distinct hepatic superior and inferior margins (Fig. 4C(vi))—achieving imaging quality comparable to clinical standards. Throughout all tasks, contact forces remained stable across all axes despite free breathing by the subjects. Notably, forces along the probe's z-axis consistently increased to approximately 3 N near standard planes, aligning with clinical practice where increased pressure enhances probe-skin contact for improved image quality. For aortic and inferior vena cava plane localization specifically, variations in z-axis torque provided valuable feedback regarding probe interactions with bony structures such as the sternum and ribs, enabling adaptive positioning strategies. In addition, movie S1 illustrates sequential acquisition of multiple standard planes on two volunteers while movie S2 presents individual standard plane localization on three subjects.

Trajectory efficiency analysis across three tasks demonstrated that our robot performed comparably to expert demonstrations, with no statistically significant differences in locating aorta planes and IVC planes (Mann-Whitney U test, P values: aorta plane = 0.509, IVC plane = 0.095), while the robot outperformed experts in portal vein plane acquisition (one-sided Mann-Whitney U test, P = 0.0006).

**Standard plane quality quantification**

Standard diagnostic images acquired by the robotic system demonstrated excellent agreement with expert-obtained images, both visually and quantitatively, across 11 subjects (six males, five females, age range: 21-53 years and BMI (body mass index) range: 18.3-24.0 kg/m²). Comparisons between robot-acquired and expert-acquired ultrasound images of four representative subjects are shown in Fig. 5. For the aorta plane (Fig. 5AB), both robotic and expert acquisitions successfully captured complete hepatic boundaries (superior, inferior, anterior, and posterior margins) with clear visualization of the aortic structure. The portal vein plane (Fig. 5C) clearly delineated the sagittal portion of the portal vein, aorta, inferior vena cava, and round ligament of the liver.

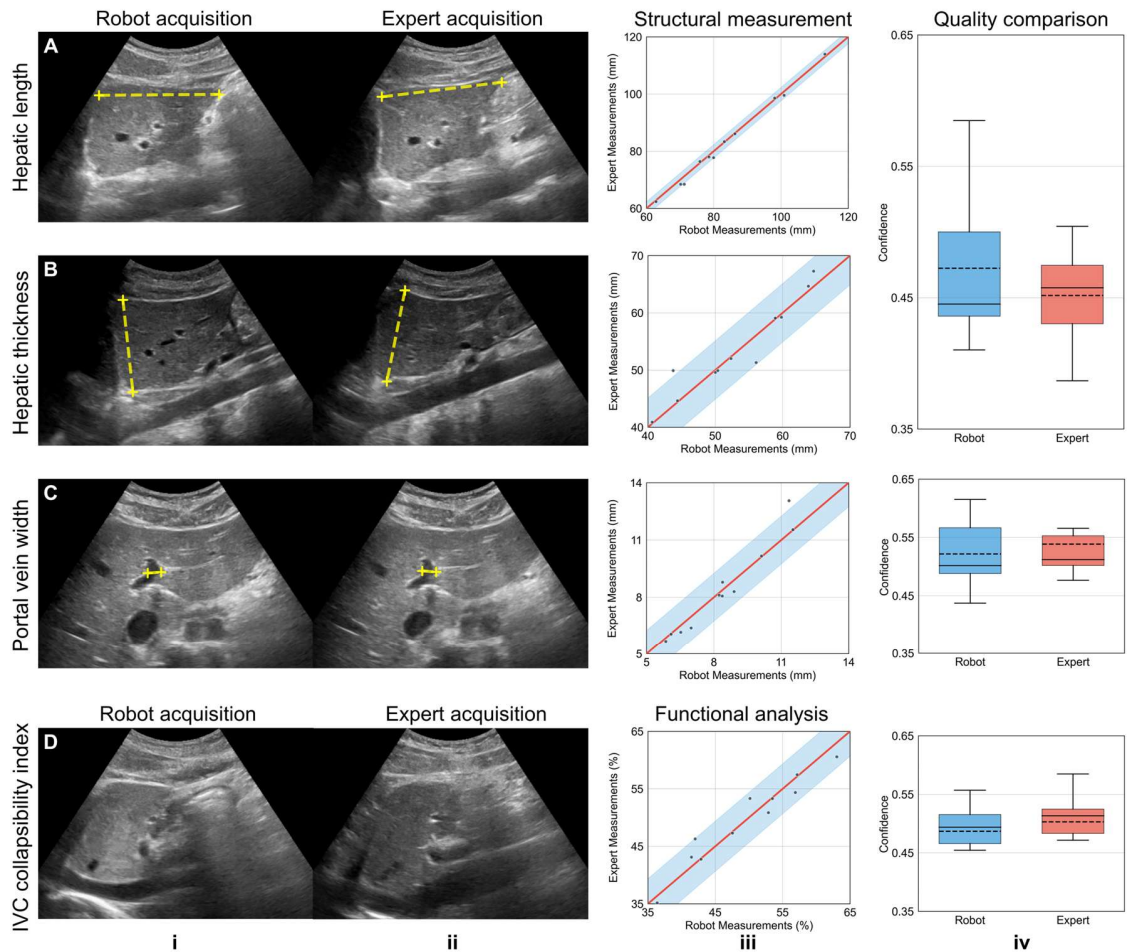

**Fig. 5. Standard plane quantification and validation.** Comparative assessment of measurement consistency between robotic and expert ultrasound acquisition on 11 volunteers. (**A**) Hepatic superior-inferior dimension measurement on standard aorta plane, (**B**) hepatic anterior-posterior dimension measurement, (**C**) portal vein width measurement on standard portal vein plane, and (**D**) inferior vena cava (IVC) collapsibility index measurement on standard IVC plane. Columns (i-ii) show representative ultrasound images acquired by the robotic system and expert sonographer, respectively, with measurement annotations. Column (iii) displays correlation analysis between expert and robotic measurements using scatter plots, where the red line represents the ideal y=x line and light blue shaded regions indicate 95% limits of agreement, demonstrating measurement concordance. Column (iv) presents image quality comparison between expert and robotic acquisitions using confidence scores (*38*) as the evaluation metric, with box plots showing median values (solid lines) and mean values (dashed lines).

Quantitative validation was performed through clinical measurements, including both anatomical structures and functional assessments. For anatomical measurements, three parameters were evaluated including hepatic length, hepatic thickness, and portal vein width.

The measurement protocols are illustrated in Fig. 5(i-ii). Strong correlations were observed between expert and robotic measurements across all anatomical parameters as demonstrated in Fig. 5A-C, iii ($R^2$ = 0.995, 0.901, and 0.950, respectively). Meanwhile, intraclass correlation coefficient (ICC) assessment revealed excellent agreement for all anatomical measurements (ICC=0.996, 0.956 and 0.965, respectively). The robotic system also demonstrated high consistency in functional assessment of the IVC collapsibility (Fig. 5D, iii, ICC = 0.966, $R^2$ = 0.929). Paired t-test confirmed no significant difference from expert assessments for the above measurements (Fig 5.A-D P values: 0.113, 0.650, 0.975, 0.918). Moreover, robotic-acquired and expert-acquired standard images showed no significant differences in the imaging quality quantified by the confidence between robotic and expert acquisitions across all three standard planes (P values: aorta plane = 0.204, portal vein plane = 0.589, IVC plane = 0.123, Fig. 5(iv)). These results demonstrate that our autonomous robotic system achieves expert-level performance in both quantitative measurements and qualitative imaging assessments.

**Autonomous pathology screening on patient**

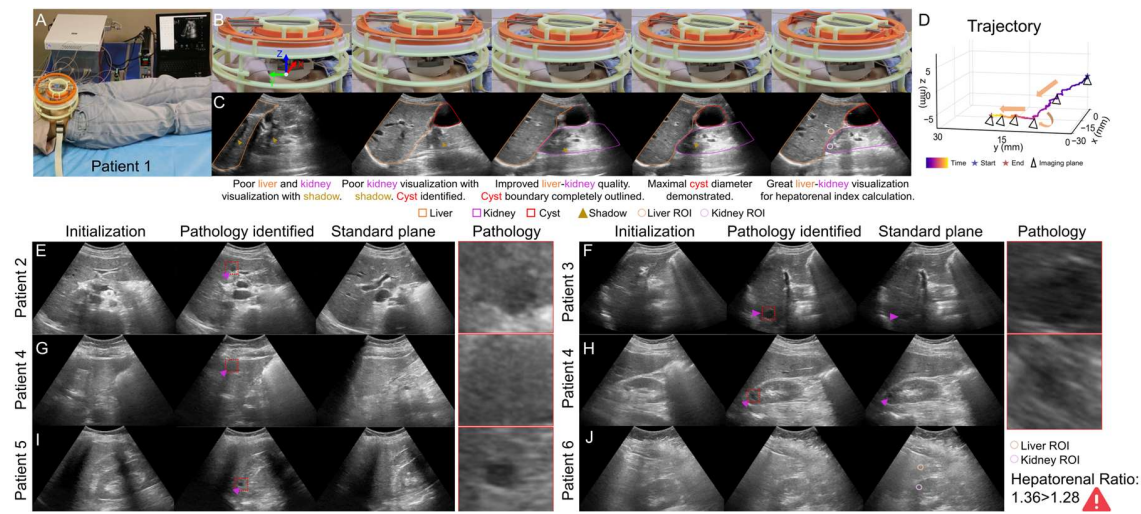

**Fig. 6. Patient evaluation.** (**A**) Experimental setup. (**B**) Sequential snapshots of robotic motion and (**C**) corresponding ultrasound image acquisition during autonomous localization of the hepatorenal standard plane in a representative patient with hepatic cyst pathology (outlined in red). (**D**) Three-dimensional motion trajectory during the autonomous examination procedure. (**E-J**) Imaging sequences including initialization frame, pathology identification frame, and the acquired standard plane, with the enlarged view of the pathology across five distinct patient cases with (E) hepatic hemangioma, (F) hepatic cyst, (G-H) hepatic and renal cysts, (I) renal cyst, and (J) suspected hepatic steatosis with abnormal hepatorenal ratio of 1.36 (normal range <1.28), indicating elevated fatty liver risk. Enlarged pathology views are provided for detailed visualization.

The diagnostic performance of the robotic system was evaluated in six patients with hepatic cysts, renal cysts, hepatic hemangioma, or fatty liver. To demonstrate the comprehensive clinical capabilities of the system, we presented a detailed case study of a 53-year-old female patient (BMI 18.3 kg/m$^2$) with a cystic lesion approximately 5 cm in diameter in the right liver. Movie S3 demonstrates the process of autonomously locating all four standard planes, measuring key anatomical structures and finding the lesion. During hepatorenal plane imaging, the robotic system effectively avoided rib interference to obtain complete cross-sectional visualization while comprehensively displaying the morphological characteristics of the cyst. Through multi-planar dynamic observation, the cyst morphology changed correspondingly with different hepatic scanning planes, consistent with typical imaging features of hepatic cysts, thereby providing reliable imaging evidence for clinical diagnosis.

Fig. 6E-J demonstrate representative screenshots from five patients (age range: 29-62 years, BMI range 17.8-29.6 kg/m$^2$) during the scanning process, with each panel comprising three sequential images of the initial image, the image containing the lesion, and the standard plane. The disease spectrum included hepatic hemangioma (Fig. 6E), hepatic cysts (Fig. 6FG), renal cyst (Fig. 6 HI), and hepatic steatosis (Fig. 6J). In the hepatic steatosis case, the hepatorenal plane revealed characteristic sonographic features including increased hepatic echogenicity with a rounded liver edge contour, diffuse parenchymal brightness exceeding renal cortical echogenicity, and posterior acoustic attenuation, with a calculated hepatorenal ratio of 1.36, indicating high risk of fatty liver disease (*39*).

**Echinococcosis screening in medically underserved regions**

The robotic ultrasound system was further deployed to Xining, Qinghai Province on the Qinghai-Tibet Plateau, the world's highest plateau, for echinococcosis screening (Fig. 7A). Compared to Beijing where the AI agent learned from experts, Xining demonstrates significantly lower medical resources. Qinghai-Tibet Plateau exhibits high endemic echinococcosis prevalence due to its pastoral characteristics, with an annual average incidence of 224.14 cases per million population, and some regions exceeding 500 cases per million population, creating a pronounced healthcare resource imbalance (Fig. 7B) (*40*). When performing autonomous sonography on a 35-year-old female (BMI: 27.3 kg/m$^2$), the system identified two echinococcosis lesions near the portal vein plane. One was located at the segment S4/S8 junction near the sagittal plane, and another in hepatic segment S7 (Fig. 7E(i)-(ii)). During aortic standard plane localization, an additional lesion was detected in hepatic segment S3 (Fig. 7E(iii)). Subsequent CT imaging performed for further diagnostic confirmation revealed three lesions consistent with the robotic ultrasound screening results (Fig. 7F). Quantitative comparison between autonomous ultrasound measurements and CT measurements demonstrated high concordance across all identified lesions (Fig. 7D), validating the system's screening potential in real-world clinical deployment scenarios within resource-limited healthcare settings. Autonomous ultrasound examination of a second patient at high risk for fatty liver disease, conducted in Xining, is presented as the second case in movie S3.

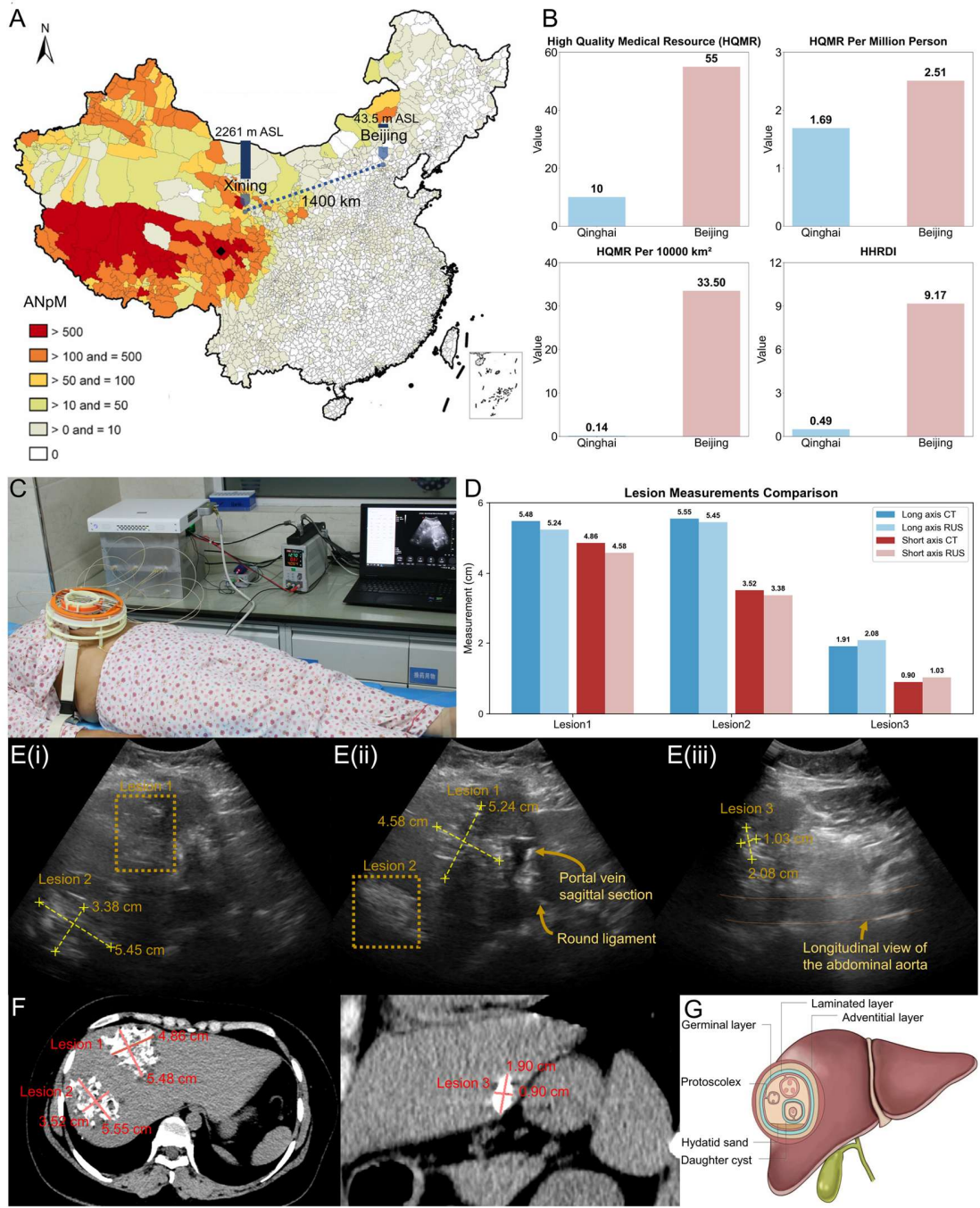

**Fig. 7. Echinococcosis screening in medically underserved regions.** (**A**) The ultrasound robot was deployed on the Qinghai-Tibet Plateau at 2261 meters above sea level and high endemic echinococcosis prevalence. (Source: Reproduced under terms of CC-BY-NC-ND license. (*3*) Copyright 2025, Wang et al, published by Elsevier.) (**B**) Comparative analysis of high-quality medical resources (HQMR) between Beijing and Qinghai Province, demonstrating significant healthcare disparities in terms of HQMR per million population, HQMR per 10,000 km², and HHRDI (High-quality Health Resource Density Index). (Source: Reproduced under terms of CC-BY license. (*40*) Copyright 2023, Yuan et al, published by Springer.) (**C**) Autonomous sonography was performed on a 35-year-old female patient with a BMI of 27.3 kg/m². (**D**) Quantitative comparison of lesion measurements between CT imaging and autonomous ultrasound assessment. (**E**) Ultrasound visualization of three identified lesions during autonomous portal vein sagittal plane (i-ii) and aortic standard plane (iii) localization, with quantitative measurements of lesion dimensions. (**F**) Corresponding CT imaging validation showing transverse and sagittal plane visualization of three lesions with dimensional measurements for evaluation only. (**G**) Schematic illustration of echinococcosis. (Source: Reproduced under terms of CC-BY-NC license. (*41*) Copyright 2023, Govindasamy et al, published by Sage.)

**Autonomous sonography under motion and in the wilderness**

The feasibility of autonomous ultrasound imaging during motion was validated on a subject walking on a treadmill. The robotic system was secured to the volunteer's abdomen via straps and autonomously performed IVC plane localization during ambulation (Fig. 8A, Movie S4). Body movement was quantified through nipple displacement, with maximum body motion reaching 14.1 cm. Abdominal flexion was quantified by measuring the distance between the nipple and the umbilicus, which varied within a range of 21.8 cm to 27.8 cm (Fig. 8C). Mechanical vibrations from the treadmill introduced ripple artifacts in the ultrasound images during motion. Despite these challenges, the robotic system successfully identified the standard IVC plane with the IVC long-axis view clearly visualized (Fig. 8D, E). These results demonstrate the robustness of our robotic system under dynamic conditions and upright postures, suggesting potential applications in emergency transport and continuous monitoring scenarios, such as IVC-based volume assessment for hemorrhage detection during patient ambulation or transport.

The portability and autonomous imaging capabilities of the robotic system in conditions where physician is hard to access were validated through ultrasound scanning in challenging wilderness conditions. The system was deployed on the grassland terrain of the Qinghai-Tibet Plateau (Fig. 8F, Movie S4), where autonomous aortic standard plane localization was performed on a healthy subject under completely uncontrolled environmental conditions. During the localization procedure, the robotic system first identified the portal vein sagittal plane and subsequently detected the aortic cross-sectional view. The system then autonomously progressed to acquire the complete aortic standard plane, achieving clear visualization of both the aortic long-axis and the superior-inferior hepatic margins (Fig. 8G). This sequential imaging progression demonstrated the system's robust performance in wilderness settings, maintaining diagnostic image quality despite environmental challenges including temperature variations and altitude effects. These multi-scenario validations collectively demonstrate the versatile, portable nature of our autonomous ultrasound system, establishing its potential for deployment in resource-limited settings, remote healthcare delivery, and emergency scenarios where traditional ultrasound expertise may be unavailable or impractical.

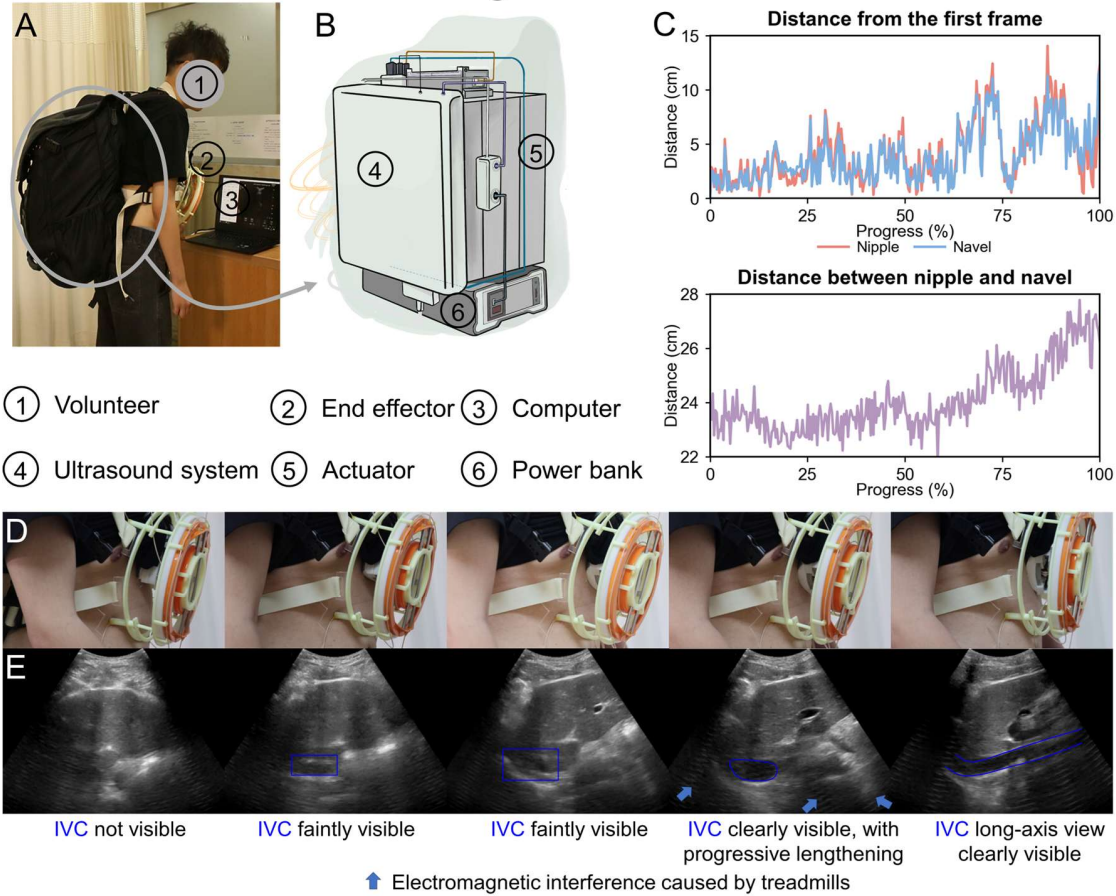

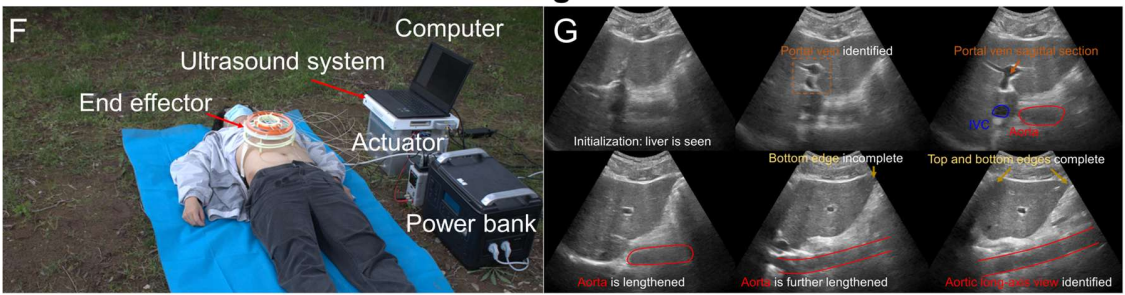

**Fig. 8. Autonomous sonography in mobile and wilderness scenarios.** (**A**) Mobile scenario experimental setup showing the robotic platform secured to a volunteer's abdomen via adjustable straps, with the subject walking on a treadmill as a dynamic condition. (**B**) Portable backpack system housing the ultrasound unit, computer, actuator, and power supply for autonomous mobile operation. (**C**) Real-time tracking of anatomical landmark displacement during subject locomotion, showing distance variations of right nipple and navel from their initial positions throughout the scanning process. Distances are calculated by combining pixel distances with hardware dimensions. Sequential visualization of (**D**) robotic motion adaptation and (**E**) corresponding ultrasound image acquisition of the IVC plane during dynamic scanning, demonstrating the system's ability to maintain imaging quality and locate key structures despite subject movement. Electromagnetic interference patterns caused by treadmill operation are indicated by blue arrows. (**F**) Wilderness deployment setup on the Qinghai-Tibet Plateau, demonstrating the system's

portability and accessibility in challenging environmental conditions. (**G**) Sequential ultrasound images during autonomous localization of the aorta plane. All key structures were successfully identified, validating the system's robust performance in the wilderness.

**DISCUSSION**

In this study, we proposed an autonomous lightweight ultrasound robot for liver sonography. We demonstrate that fusing image, pose and force features and utilizing memory attention mechanism to process sequential inputs represent an effective approach for locating unseen structure from two-dimensional sequences. Through imitation learning from expert, the system autonomously performed the complex and non-continuous liver sonography for structural measurements, functional evaluation, and lesion detection, and achieved expert-level imaging quality and accuracy. Furthermore, benefiting from the lightweight design and abdominal mounting configuration, our robot exhibits exceptional adaptability across diverse clinical scenarios. Beyond traditional hospital-based diagnostics, our system can be effectively deployed in challenging environments, including dynamic settings and remote wilderness environments.

Our ultrasound robot demonstrates superior generalization performance across diverse subjects and scenarios. The training and testing cohorts were completely separated, with the testing population including patients with pathological conditions that introduced out-of-distribution challenges not encountered during the training phase. We attribute this robust generalization capability to the domain randomization. While domain randomization is predominantly employed in reinforcement learning for sim-to-real transfer, it proves equally beneficial for imitation learning when addressing distributional shifts. By deliberately obscuring structure-irrelevant visual features during training, this approach compels the AI agent focus exclusively on task-relevant anatomical landmarks and operations. This selective attention mechanism enables the AI agent to maintain stable performance when encountering novel visual presentations, pathological variations, or environmental conditions that deviate from the training distribution.

This study demonstrates the potential of imitation learning-based autonomous medical robots to address critical healthcare gaps in resource-limited settings. Unlike physician training requiring years of clinical practice, imitation learning achieves expert-level proficiency within hours. Our algorithm has excellent scalability, as additional ultrasound examination protocols can be integrated by simply collecting more expert demonstrations without modifying the data acquisition pipeline or model architecture. Through our demonstration of deploying the robotic system on the Qinghai-Tibet Plateau for hepatic echinococcosis screening, we show that in emergency medicine, wilderness medicine, and medically underserved regions where specialist physicians are unavailable, autonomous medical robots equipped with expert-level AI agent can deliver sophisticated diagnostic interventions that would otherwise be inaccessible, thereby extending high-quality medical services to previously unreachable populations and clinical scenarios.

The abdominal-mounted robotic paradigm presents significant potential for advancing abdominal medical robotics design. This approach eliminates the relative motion risks inherent in traditional standalone robotic arm systems by establishing direct mechanical coupling with the human body at the structural level. The superior coupling between our robotic system and the human body enables real-time compensation for patient respiratory motion and involuntary body movements and ensures stable relative positioning between the probe and anatomical structures throughout autonomous sonography. These reduce the technical complexity of abdominal ultrasound scanning. Meanwhile, this human-robot-integrated design provides inherent mechanical stability and safety advantages in dynamic medical environments, such as medical transport scenarios, where traditional rigid robotic arm systems may generate hazardous impact loads or unexpected disconnections due to sudden acceleration, deceleration, or vehicle vibrations. Furthermore, this human-coupled ultrasound robot surpasses existing wearable ultrasound sensors, which are limited to superficial tissues (typically <5cm) with relatively fixed detection targets. In contrast, our robot enables flexible and autonomous switching between different imaging planes during human locomotion while maintaining clinical-grade imaging depth of 18cm, providing unprecedented technical means for dynamic abdominal organ assessment in exercise physiology and real-time anatomical structure monitoring. The ability to perform autonomous, high-quality abdominal imaging during patient movement represents a paradigm shift in point-of-care ultrasound, potentially enabling new diagnostic workflows in emergency medicine, sports medicine, and continuous patient monitoring scenarios where traditional sonography by a doctor is impractical or difficult.

Furthermore, our system exhibits remarkable deployment flexibility due to the coordinate-system-agnostic framework. The AI agent requires no calibration between the coordinate systems of the expert demonstration acquisition setup and the deployment robotic platform. This capability enables seamless deployment across 6-degree-of-freedom robotic manipulators, fundamentally decoupling the AI agent from specific hardware configurations. These combined capabilities enable a paradigm of centralized expertise acquisition with distributed deployment scalability. Expert demonstrations can be collected at a single specialized center and subsequently deployed across multiple clinical sites with heterogeneous robotic platforms.

We also see several limitations. First, our system relies on commercial or commercially-derived ultrasound probes and force sensors, which contribute to size and weight constraints that could be addressed through novel hardware. Second, although we successfully achieved deep organ imaging during dynamic conditions, the clinical utility and diagnostic potential of this approach warrant additional study.

# MATERIALS AND METHODS

## Agent architecture

### *Pose harmonization*

A relative pose representation that harmonized pose across different coordinate frames was adopted to enhance the generalization capability of the AI agent across different coordinate frames, including the acquisition system and different robot platforms. The large variations in global position and orientation were normalized into consistent, small-scale relative movements. Given two probe poses in the world frame $\mathbf{P}_{1,world} = \begin{bmatrix} \mathbf{R}_{1,world} & \mathbf{t}_{1,world} \\ 0 & 1 \end{bmatrix}$ and $\mathbf{P}_{2,world} = \begin{bmatrix} \mathbf{R}_{2,world} & \mathbf{t}_{2,world} \\ 0 & 1 \end{bmatrix}$ ( $\mathbf{R}_{i,coord} \in \mathbb{R}^{3\times3}$ , $\mathbf{t}_{i,coord} \in \mathbb{R}^{3\times1}$ ), the relative transformation of $\mathbf{P}_{2,world}$ with respect to the local frame of $\mathbf{P}_{1,world}$ was

$$\mathbf{T}_{\mathbf{P}_1 \to \mathbf{P}_2, \mathbf{P}_1} = RelP(\mathbf{P}_{2,world}, \mathbf{P}_{1,world})$$

$$= \begin{bmatrix} \mathbf{R}_{1,world}^T \mathbf{R}_{2,world} & \mathbf{R}_{1,world}^T (\mathbf{t}_{2,world} - \mathbf{t}_{1,world}) \\ 0 & 1 \end{bmatrix} \quad (1)$$

For demonstration ending at time *t*, all poses were transformed into the local frame of the terminal pose before network input. For robot deployment, the inverse operator $\boldsymbol{RelPInv}(\cdot,\cdot)$ mapped relative pose sequences to any 6-DoF robotic platform. Given the current TCP pose $\mathbf{P}_{\text{TCP,robot}}$ in the robot frame, the transformation matrix of action was

$$\mathbf{T}_{\mathbf{P}_1 \to \mathbf{P}_2,\text{robot}} = \boldsymbol{RelPInv}(\mathbf{T}_{\mathbf{P}_1 \to \mathbf{P}_2,\mathbf{P}_1}, \mathbf{P}_{\text{TCP,robot}}) = \mathbf{P}_{\text{TCP,robot}} \cdot \mathbf{T}_{\mathbf{P}_1 \to \mathbf{P}_2,\mathbf{P}_1} \cdot \mathbf{P}_{\text{TCP,robot}}^{-1}. \quad (2)$$

*Image randomization*

Domain randomization was applied to ultrasound images to enhance the generalization capability of the policy network across different individuals and imaging settings. The randomization strategy considered common ultrasound imaging artifacts, incorporating Gaussian noise with standard deviation ranging from 0.01 to 0.1, speckle noise with intensity ranging from 0 to 0.3 and beam pattern effects with intensity between 0.3 to 0.7. Additionally, image contrast and brightness were randomly perturbed within the range of 0.7 to 1.3 and 0.7 to 1.3 to simulate ultrasound imaging under varying adipose tissue conditions. Random deformation with intensity from 0.15 to 0.5 was applied to liver structures to simulate anatomical variations across different individuals.

*Multi-modal feature fusion*

The AI agent processed temporal observation sequences of length *T*, comprising ultrasound images $[I_1, \dots, I_T]$, six-dimensional wrench measurements $[F_1, \dots, F_T]$, and probe poses $[P_1, \dots, P_T]$. All poses were transformed to the reference frame of the terminal pose $P_T$, yielding harmonized poses $[P_{1|T}, \dots, P_{T|T}]$. Each ultrasound image was processed through the encoder of ResNet-18 and flattened into a feature vector. For pose and force data, we employed Fourier positional encoding to map continuous pose and force values to a high-dimensional frequency domain, enhancing the modeling of multi-scale motion patterns and periodic behaviors while preserving spatial-temporal continuity essential for precise manipulation:

$$\gamma(p) = [\sin(2^0 \pi p), \cos(2^0 \pi p), \dots, \sin(2^{L-1} \pi p), \cos(2^{L-1} \pi p)]. \quad (3)$$

A one-layer Kolmogorov-Arnold network (KAN) then projected all features to the same number of dimensions.

The cross-attention between pose feature and image feature was proposed to identify key frames of long state sequence as memory. The cross-attention weights between pose queries and image keys were computed as:

$$\boldsymbol{A}_t = \text{softmax}\left(\frac{\boldsymbol{Q}_{pose}(P_t) \cdot \boldsymbol{K}_{image}(I_t)^T}{\sqrt{d_k}}\right) \quad (4)$$

where $\boldsymbol{Q}_{pose}(\cdot)$ and $\boldsymbol{K}_{image}(\cdot)$ were learned query and key projection functions. To capture task-critical information while maintaining computational efficiency, *k*=5 frames with highest cumulative attention weights were selected as representative keyframes, effectively compressing redundant observations.

Based on the identified keyframes, cross-attention features were computed between image and pose. In parallel, encoded forces and poses of the keyframes were mapped through a one-layer KAN to produce task-aware tactile features. The AI agent finally conducted multi-modal fusion between image-pose and force-pose features using a two-layer KAN, effectively capturing the intricate coupling between probe kinematics and contact dynamics essential for precise ultrasound operation.

*Multi-dimensional action generalization*

A conditional diffusion model (*42*) was trained to generate multi-step action trajectories, where each action comprised a 6-DoF target pose and six-dimensional force vector. Given a demonstration trajectory $\boldsymbol{D}_0 = [\boldsymbol{\tau}_0^T \ \boldsymbol{\tau}_1^T \ \cdots \ \boldsymbol{\tau}_K^T]^T$, where $\boldsymbol{\tau}_i^T = [\mathbf{T}_{P_0 \to P_i, P_0}^T \ \mathbf{F}_i^T]$ represented the pose transformation and force at step *i*, the forward diffusion process progressively corrupted the trajectory through Gaussian noise injection:

$$\boldsymbol{D}_t = \sqrt{\alpha_t}\boldsymbol{D}_{t-1} + \sqrt{1-\alpha_t}\boldsymbol{\epsilon}, \boldsymbol{\epsilon} \sim N(\mathbf{0}, \mathbf{I}). \qquad (5)$$

The denoising network $\boldsymbol{De}(\cdot)$ learned to reverse this process by predicting the original trajectory from its noised version, conditioned on fused image-pose-force features $\boldsymbol{C}$. Training optimized the following loss:

$$\text{Loss} = \boldsymbol{MSE}(\boldsymbol{De}(\boldsymbol{\tau}_t; t, \boldsymbol{C}), \boldsymbol{\tau}_0). \qquad (6)$$

During inference, executable trajectories $\hat{\boldsymbol{\tau}}_0$ were generated from random Gaussian noise through iterative denoising following the reverse sampling process:

$$\hat{\boldsymbol{\tau}}_{t-1} = \frac{(1-\overline{\alpha}_{t-1})\sqrt{\alpha_t}}{1-\overline{\alpha}_t}\hat{\boldsymbol{\tau}}_t + \frac{(1-\alpha_t)\sqrt{\overline{\alpha}_{t-1}}}{1-\overline{\alpha}_t}\boldsymbol{De}(\hat{\boldsymbol{\tau}}_t; t, \boldsymbol{C}) + \boldsymbol{\sigma}_q(t)\boldsymbol{\epsilon}, \boldsymbol{\epsilon} \sim N(\mathbf{0}, \mathbf{I}) \qquad (7)$$

where $\overline{\alpha}_t$, $\alpha_t$, and $\boldsymbol{\sigma}_q(t)$ were time-step dependent variance scheduling functions that controlled the noise injection and sampling process respectively. All models were trained for 200 epochs.

**Data acquisition system**

A synchronized data acquisition system was developed to capture comprehensive manipulation data during expert ultrasound scanning procedures. The system comprised three integrated components: a compact convex ultrasound probe (UD32, Hisky, China), an OptiTrack (V120-Trio, OptiTrack, America) optical tracking system, and a six-axis force sensor (γ45, Daysensor, China). The probe was directly mounted onto the force sensor to ensure accurate measurement of contact forces during probe-tissue interaction, while a handheld grip was attached to the opposite end of the force sensor to facilitate natural manipulation. Optical tracking markers were positioned on the grip assembly to maintain continuous visibility throughout the scanning procedure, enabling precise 6-DoF pose tracking without occlusion during expert demonstrations.

**Hardware implementation**

The end-effector was fabricated via stereolithography 3D printing using 8228 resin, providing high mechanical strength and durability. For translational degrees of freedom, stainless steel rods (diameter: 5 mm) served as linear guideways (Fig. 3C), while a precision glass bearing (outer diameter: 210.5 mm, inner diameter: 190.5 mm) supported the revolute joint (Fig. 3D). Vertical actuation was realized through a cable-actuated compression mechanism incorporating medical-grade superelastic NiTi (shape memory alloy) springs housed within the support legs for controlled height adjustment (Fig. 3E). For rotational degrees of freedom, the three-axis gimbal structure was machined from aluminum alloy to minimize cable friction and enhance rotational control precision for roll and pitch (Fig. 3FG). A larger glass bearing (outer diameter: 108.9 mm, inner diameter: 88.9 mm) enabled rotation in Fig. 3H.

The actuation system comprised six NEMA 17 stepper motors, each driving individual Ultra High Molecular Weight Polyethylene cable to control the respective degrees of freedom. A mechanical tensioning mechanism employs a steel rod to generate constant preload force (approximately 2.5 N) on one side of each cable, maintaining baseline tension throughout the operational range. Cable routing utilized low-friction, self-lubricating PEEK (polyether ether ketone) tubing to preserve tensile loading and minimize wear. Movie S5 illustrates the detailed robotic system design and degrees of freedom.

**Ethical compliance**

All experimental procedures involving human participants in this study were conducted in accordance with the Declaration of Helsinki and approved by Tsinghua University Science and Technology Ethics Committee (THU-01-2025-0088, THU-01-2025-1006). Written informed consent was obtained from all volunteers prior to participation.

**Human experimental setup**

Demonstration data were collected by an expert sonographer with more than 10 years of clinical experience. For the aorta plane, IVC plane and portal vein plane, the ultrasound probe was initially positioned randomly within the subxiphoid abdominal region, while for the hepatorenal plane, initial placement was randomized along the right subcostal margin. The acquisition system recorded complete trajectories from initial plane to the target plane. A total of 134, 140, 184, and 183 demonstration sequences were recorded, yielding 51336, 50301, 66685, and 18378 image-pose-force data pairs for the four tasks, respectively.

The trained AI agent was subsequently deployed on the robotic system for clinical validation on an independent test cohort with complete separation from the training population. The robotic system was mounted on the volunteers via securing straps, maintaining consistent relative position between the robot and volunteer.

**Phantom experimental setup**

Phantom experiments were conducted on a standard abdominal phantom model (Model 057A, CIRS, USA) using a commercial UR3e robotic arm (Universal Robots, Denmark) to enable pure algorithmic comparison by eliminating hardware-related confounding factors. Three standard planes corresponding to the aorta plane, IVC plane and portal vein plane in the human experiment were defined in the phantom. Additionally, these planes contained one, two and four tumors, respectively. The ground truth poses were recorded for each plane. For demonstration data acquisition, the ultrasound probe was initially positioned randomly on the phantom. The acquisition system recorded expert manipulation trajectories from initial positioning to successful target plane acquisition, capturing 46, 43, and 43 demonstration sequences for the three respective tasks, yielding 5649, 6151, and 4019 image-pose-force data pairs.

The proposed method was compared against a diffusion policy baseline. The baseline implementation followed the official diffusion policy framework with task-specific adaptations: image features, probe poses, and contact forces were concatenated as conditional inputs to the diffusion model, while the network outputs predicted both poses and forces. During evaluation, both methods were tested from 14 starting positions across the phantom surface to locate the standard planes.

**Statistical analysis**

Statistical analysis was conducted using Python (version 3.9.23), NumPy (version 1.25.2) and SciPy (version 1.11.2). The number of samples were reported in the figure caption. *P* values were reported. A *P* value smaller than 0.05 was considered statistically significant.

**Supplementary Materials**
 Fig. S1
 Table S1
 Movies S1 to S5

**Supplementary Materials**

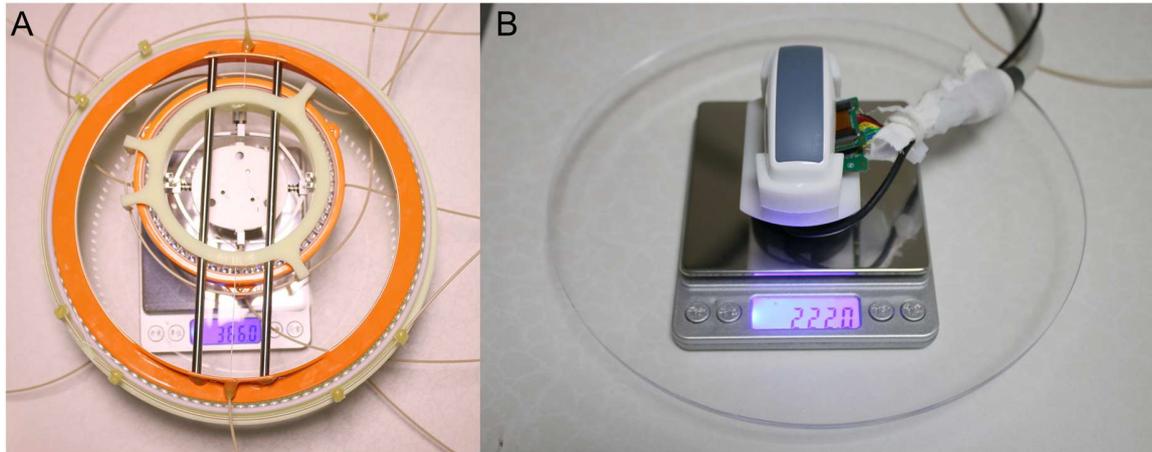

**Fig. S1. The weight of the lightweight ultrasound robot.** The end effector weights 366.0 grams while the ultrasound probe and the force sensor weight 222.0 grams.

**Table S1. Performance parameters of the lightweight ultrasound robot.** Motion accuracy and range for each degree of freedom. Accuracy is quantified as the mean absolute error between theoretical and measured position values across eight experiments for each degree of freedom. "Regulator" indicates results with cable tension regulator enabled.

|  | $l$ (mm) | $\theta$ (°) | $z$ (mm) | $\alpha$ (°) | $\beta$ (°) | $\gamma$ (°) |
|---|---|---|---|---|---|---|
| Accuracy (with regulator) | 0.46 | 0.60 | \ | 0.56 | 1.37 | 1.33 |
| Accuracy (w/o regulator) | 0.51 | 0.95 | 1.00 | 2.11 | 2.89 | 2.50 |
| Range | -16~+16 | -100~100 | -8~4 | -15~15 | -25~25 | -100~100 |